\documentclass[%
 reprint,
 amsmath,amssymb,
prb,
]{revtex4-2}

\usepackage{lipsum}
\usepackage{bm}
\usepackage{amssymb}
\usepackage{amstext}
\usepackage{amsmath}
\usepackage{mathrsfs}
\usepackage{graphicx}
\usepackage{color}
\usepackage{amsthm}
\usepackage{floatrow}
\usepackage{array}
\renewcommand\arraystretch{2}
\usepackage{listings}
\usepackage{booktabs}

\definecolor{codegreen}{rgb}{0,0.6,0.4}
\definecolor{codegray}{rgb}{0.5,0.5,0.5}
\definecolor{codepurple}{rgb}{0.58,0,0.82}
\definecolor{backcolour}{rgb}{0.97,0.97,0.97}
\definecolor{keywordcolor}{rgb}{0,0,0.8}

\lstdefinestyle{mystyle}{
    backgroundcolor=\color{backcolour},   
    commentstyle=\color{codegreen},
    keywordstyle=\color{keywordcolor},
    numberstyle=\tiny\color{codegray},
    stringstyle=\color{codepurple},
    basicstyle=\ttfamily\footnotesize,
    breakatwhitespace=false,         
    breaklines=true,                 
    captionpos=b,                    
    keepspaces=true,                 
    numbers=left,                    
    numbersep=5pt,                  
    showspaces=false,                
    showstringspaces=false,
    showtabs=false,                  
    tabsize=2
}
\usepackage{hyperref}

\begin{document}
\preprint{APS/123-QED}
\title{Koopman neural operator as a mesh-free solver of non-linear partial differential equations}
\thanks{Correspondence should be addressed to X.R.D., P.S., and Y.T.}%

\author{Wei Xiong}
\altaffiliation{Department of Earth System Science, Tsinghua University, Beijing, 100084, China.}

 \author{Xiaomeng Huang}
 \altaffiliation{Department of Earth System Science, Tsinghua University, Beijing, 100084, China.}
 
 \author{Ziyang Zhang}
 \altaffiliation{Markov Laboratory, Central Research Institute, 2012 Laboratories, Huawei Technologies Co. Ltd., China.}
 
\author{Ruixuan Deng}%
\email{ruiton@umich.edu}
 \altaffiliation{Department of Statistics, University of Michigan, Ann Arbor, MI 48104, Michigan, United States of America.}
 
\author{Pei Sun}%
\email{peisun@cityu.edu.mo}
 \altaffiliation{Faculty of Health Sciences, City University of Macau, Macau, 999078, China.}

 \author{Yang Tian}
\email{tyanyang04@gmail.com}
 \altaffiliation{Faculty of Health Sciences, City University of Macau, Macau, 999078, China.}
 \altaffiliation[Also at ]{Faculty of Data Science, City University of Macau, Macau, 999078, China.}
 



\begin{abstract}
The lacking of analytic solutions of diverse partial differential equations (PDEs) gives birth to a series of computational techniques for numerical solutions. Although numerous latest advances are accomplished in developing neural operators, a kind of neural-network-based PDE solver, these solvers become less accurate and explainable while learning long-term behaviors of non-linear PDE families. In this paper, we propose the Koopman neural operator (KNO), a new neural operator, to overcome these challenges. With the same objective of learning an infinite-dimensional mapping between Banach spaces that serves as the solution operator of the target PDE family, our approach differs from existing models by formulating a non-linear dynamic system of equation solution. By approximating the Koopman operator, an infinite-dimensional operator governing all possible observations of the dynamic system, to act on the flow mapping of the dynamic system, we can equivalently learn the solution of a non-linear PDE family by solving simple linear prediction problems. We validate the KNO in mesh-independent, long-term, and5zero-shot predictions on five representative PDEs (e.g., the Navier-Stokes equation and the Rayleigh-B{\'e}nard convection) and three real dynamic systems (e.g., global water vapor patterns and western boundary currents). In these experiments, the KNO exhibits notable advantages compared with previous state-of-the-art models, suggesting the potential of the KNO in supporting diverse science and engineering applications (e.g., PDE solving, turbulence modelling, and precipitation forecasting). 

\end{abstract}

\maketitle
\section{Introduction}
\subsection{Partial differential equation solvers are important}
In science and engineering, partial differential equation (PDE) is a fundamental tool to characterize various problems (e.g., problems in fluid mechanics, quantum mechanics, and civil engineering) \cite{debnath2005nonlinear}. However, even though significant progress has been achieved on solving PDEs \cite{tanabe2017functional}, numerous important PDEs, such as the Navier-Stokes equation, still lack analytic solutions \cite{gockenbach2005partial}. Consequently, the urgent needs of solving complicated PDEs in real applications have given birth to diverse techniques for computationally approximating PDE solutions \cite{mattheij2005partial}. 

Given $\Phi=\Phi\left(D;\mathbb{R}^{d_{\phi}}\right)$, a Banach space of inputs, and $\Gamma=\Gamma\left(D;\mathbb{R}^{d_{\gamma}}\right)$, a Banach space of solutions, defined on a bounded open set $D\subset\mathbb{R}^{d}$, these PDE solvers are expected to approximate a solution operator $\mathcal{Q}$ that relates $\Phi$ with $\Gamma$ for a typically time-dependent PDE family
\begin{align}
\partial_{t}\gamma\left(x_{t}\right)&=\left(\mathcal{L}_{\phi}\gamma\right)\left(x_{t}\right)+\eta\left(x_{t}\right),\;x_{t}\in D\times T,\label{EQ1}\\
\gamma\left(x_{t}\right)&=\gamma_{B},\;x_{t}\in\partial D\times T,\label{EQ2}\\
\gamma\left(x_{0}\right)&=\gamma_{I},\;x_{0}\in D\times \{0\}.\label{EQ3}
\end{align}
In Eq. (\ref{EQ1}-\ref{EQ3}), notions $\gamma_{B}$ and $\gamma_{I}$ denote the boundary and initial conditions, set $T=\left[0,\infty\right)$ is the time domain, differential operator $\mathcal{L}_{\phi}$ is characterized depending on $\phi$, fixed function $\eta\left(\cdot\right)$ lives in an appropriate function space determined by $\mathcal{L}_{\phi}$, and $\gamma\left(\cdot\right)$ is the solution of the PDE family. This formulation gives rise to the solution operator $\mathcal{Q}:\left(\phi,\gamma_{B},\gamma_{I}\right)\mapsto\gamma$, which reduces to $\mathcal{Q}:\phi\mapsto\gamma$ if boundary and initial conditions are constant. For convenience, we always consider fixed boundary and initial conditions in our subsequent derivations. In application, researchers consider a parametric counterpart $\mathcal{Q}_{\theta}\simeq \mathcal{Q}$ parameterized by $\theta$ to define optimization problems \cite{li2020neural}. 

\subsection{Existing partial differential equation solvers are diverse} To date, diverse types of PDE solvers have been developed, which can be generally classified as following:
\begin{itemize}
    \item[(1) ] \textbf{Classic numerical solvers.} The earlist and commonest PDE solvers, such as finite element \cite{reddy2019introduction} and finite difference \cite{lipnikov2014mimetic} methods, solve PDEs by discretizing the space according to specific mesh designs. These approaches are granularity-dependent, whose accuracy favors fine-grained meshes while efficiency prefers coarse-grained meshes \cite{tadmor2012review}. Therefore, they inevitably face the tradeoff between accuracy and efficiency when the target PDE is complicated \cite{li2020fourier}.
    \item[(2) ] \textbf{Neural-network-based solvers.} To revolutionize the computational techniques of PDE solving, three types of neural-network-based solvers have been proposed to approximate or enhance the classic ones in a fast manner \cite{raissi2019physics,kochkov2021machine}:
    \begin{itemize}
        \item[(2a) ] \textbf{Mesh-dependent and finite-dimensional operators.} The first type of solvers approximate the solution operator as a parameterized neural network between finite Euclidean spaces after discretizing domains $D$ and $T$ into $x$ and $y$ meshes, i.e., $\mathcal{Q}_{\theta}:\mathbb{R}^{x}\times\mathbb{R}^{y}\times\Theta\rightarrow\mathbb{R}^{x}\times\mathbb{R}^{y}$ \cite{guo2016convolutional,zhu2018bayesian,bhatnagar2019prediction}. These solvers are mesh-dependent and require fine-tuning on different values of $n$, leading to limited generalization capacities \cite{li2020neural}. 
        \item[(2b) ] \textbf{Neural finite element methods.} The second type of solvers directly parameterize equation solution $\gamma\left(\cdot\right)$ as a neural network, which equivalently gives rise to $\mathcal{Q}_{\theta}:D\times T\times\Theta\rightarrow\mathbb{R}$ \cite{yu2018deep,raissi2019physics,bar2019unsupervised,pan2020physics,lienen2022learning}. Although these solvers are mesh-independent and accurate, they are limited to learn a certain instance of the PDE rather than the entire family \cite{li2020neural}. Therefore, similar to the classic numerical ones, these solvers requires new network design and training whenever the instance is changed. Meanwhile, most of these solvers are not applicable to the cases where the underlying PDE remains unknown (see an exception in the finite element network \cite{lienen2022learning}, which supports spatio-temporal forecasting on real data). 
        \item[(2c) ] \textbf{Neural operators.} The third type of solvers are developed to learn a mesh-dependent and infinite-dimensional solution operator with neural networks, i.e, $\mathcal{Q}_{\theta}:\Phi\times\Theta\rightarrow\Gamma$ \cite{lu2019deeponet,bhattacharya2020model,nelsen2021random,li2020neural,li2020fourier,kovachki2021universal,li2022fourier}. These solvers overcome the dependence on meshes by learning network parameters in a manner applicable to different discretizations. Because these solvers learn the solution operator directly, they only need to be trained once for a target PDE family. Generating equation solution $\gamma\left(\cdot\right)$ of different instances of the PDE family only requires a forward pass of networks, which is computationally favorable \cite{li2020neural,li2020fourier}. Although neural operators are initially not competitive with other neural-network-based solvers because evaluating kernal integral operators is costly, the latest approach, named as Fourier neural operator \cite{li2020fourier}, resolves this limitation by fast Fourier transform.
    \end{itemize}
\end{itemize}

Compared with classic numerical solvers, neural-network-based solvers, especially neural operators, are more efficient in dealing with science and engineering questions where PDEs are complicated \cite{li2020neural,li2020fourier}. Therefore, our research primarily focus on neural operator designs.

\subsection{Existing partial differential equation solvers face challenges} Despite substantial progress achieved by neural operators in theoretical foundations (e.g., Ref. \cite{kovachki2021universal}), approximator designs (e.g., Refs. \cite{li2020fourier,li2022fourier}), and applications (e.g., Ref. \cite{guibas2021efficient}), there still remain numerous challenges in existing neural operator solvers, among which, a critical one lies in the limited capacity of existing models to learn the long-term dynamics of complicated PDEs. 

To understand this challenge, let us consider a Green function, $\mathcal{J}_{\phi}:\left(D\times T\right)\times \left(D\times T\right)\rightarrow \mathbb{R}$, associated with Eqs. (\ref{EQ1}-\ref{EQ3}) \cite{li2020neural}
\begin{align}
&\gamma\left(x_{t+\varepsilon}\right)=\int_{D\times \{t\}}\mathcal{J}_{\phi}\left(x_{t},y_{t}\right)\eta\left(y_{t}\right)\mathsf{d}y_{t},\;\forall \;x_{t}\in D\times \{t\}.\label{EQ4}
\end{align}
The initial idea of neural operators \cite{li2020neural} is to parameterize this Green function as a kernel integral operator (i.e., see the integral term related to $\kappa_{\theta}$ presented below) and define an iterative update strategy of neural networks
\begin{widetext}
  \begin{align}
&\widehat{\gamma}\left(x_{t+\varepsilon}\right)=\sigma\left(W\widehat{\gamma}\left(x_{t}\right)+\int_{D\times \{t\}}\kappa_{\theta}\left(x_{t},y_{t},\phi\left(x_{t}\right),\phi\left(y_{t}\right)\right)\widehat{\gamma}\left(y_{t}\right)\mathsf{d}y_{t}\right),\;\forall \;x_{t}\in D\times \{t\},\label{EQ5}
\end{align}  
\end{widetext}
where $\varepsilon\in\left(0,\infty\right)$ denotes time difference, notion $\widehat{\gamma}:D\times T\rightarrow\mathbb{R}^{d_{\widehat{\gamma}}}$ denotes the neural network representation of equation solution $\gamma$ generated by specific embedding, mapping $\sigma:\mathbb{R}\rightarrow\mathbb{R}$ denotes an element-wise activation function, notion $W:\mathbb{R}^{d_{\widehat{\gamma}}}\rightarrow\mathbb{R}^{d_{\widehat{\gamma}}}$ denotes a linear transformation, and $\kappa_{\theta}:\mathbb{R}^{2\left(d+d_{\phi}\right)}\rightarrow\mathbb{R}^{d_{\widehat{\gamma}}}$ is a neural network parameterized by $\theta$ \cite{li2020neural}. The computing efficiency of this iterative update can be improved using the well-known Fourier neural operator \cite{li2020fourier}, i.e., Eq. (\ref{EQ5}) can be reformulated as $\widehat{\gamma}\left(x_{t+\varepsilon}\right)=\sigma\left[W\widehat{\gamma}\left(x_{t}\right)+\mathcal{F}^{-1}\left(\mathcal{F}\left(\kappa_{\theta}\right)\cdot\mathcal{F}\left(\widehat{\gamma}\left(x_{t}\right)\right)\right)\right]$ for any $x_{t}\in D\times \{t\}$ (notion $\mathcal{F}\left(\cdot\right)$ is the Fourier transform that can be realized by fast Fourier transform in application). 

At the first glance, the iterative update in Eq. (\ref{EQ5}) is similar to a dynamic system perspective where we study the evolution mapping $\zeta:\mathbb{R}^{d_{\gamma}}\times T\rightarrow\mathbb{R}^{d_{\gamma}}$ of an infinite-dimensional non-linear dynamic system $\gamma_{t}=\gamma\left(D\times\{t\}\right)$ (notion $\gamma\left(D\times\{t\}\right)$ represents that function $\gamma$ acts on all elements in set $D\times\{t\}$)
\begin{align}
\gamma_{t+\varepsilon}=\gamma_{t}+\int_{t}^{t+\varepsilon}\zeta\left(\gamma_{\tau},\tau\right)\mathsf{d}\tau,\;\forall t\in T.\label{EQ6}
\end{align}
In this dynamic system, each snapshot $\gamma_{t}$ is a field distributed on domain $D$ at moment $t$, i.e., a time-dependent equation solution of Eqs. (\ref{EQ1}-\ref{EQ3}). The temporal evolution of field $\gamma_{t}$ is characterized by an infinite-dimensional evolution mapping $\zeta:\mathbb{R}^{d_{\gamma}}\times T\rightarrow\mathbb{R}^{d_{\gamma}}$
\begin{align}
\partial_{t}\gamma_{t}=\zeta\left(\gamma_{t},t\right),\;\forall\gamma_{t}\in\mathbb{R}^{d_{\gamma}}\times T,\label{EQ7}
\end{align}
which lies in the heart of dynamic system theories \cite{perko2013differential,fishwick2007handbook}. When domain $D$ stands for a bounded spatial range, Eq. (\ref{EQ5}) can be understood as a general spatio-temporal dynamics model of diverse phenomena in neuroscience \cite{bressloff2011spatiotemporal}, geophysics \cite{herrera2017spatiotemporal}, fluid mechanics \cite{wu2020spatio}, and engineering \cite{scholl2001nonlinear}.

For most non-linear PDE families, the concerned evolution mapping $\zeta\left(\cdot,\cdot\right)$ is even more intricate than the equation solution $\gamma\left(\cdot\right)$ itself. In the terminologies of modern dynamic system theory \cite{brunton2022modern}, Eq. (\ref{EQ6}) generally leads to the cocycle property of the flow mapping $\theta$
\begin{align}
\theta_{t}^{t^{\prime}}=\theta_{t+\tau}^{t^{\prime}}\circ\theta_{t}^{t+\tau},\;\forall t\leq t+\tau\leq t^{\prime}\in T\label{EQ7}
\end{align}
where notion $\circ$ denotes the composition of mappings. This cocycle property defines how $\gamma\left(\cdot\right)$, the equation solution of target PDE family, evolves across adjoining time intervals. Challenges frequently arise as time difference $\varepsilon$ in Eq. (\ref{EQ6}) enlarges because the dynamic system of $\gamma\left(\cdot\right)$ can be either non-autonomous (i.e., the flow mapping $\theta$ associated with $\zeta\left(\cdot,\cdot\right)$ is time-dependent) or autonomous (i.e., the flow mapping $\theta$ is time-independent such that $\partial_{t}\zeta\left(\cdot,t\right)\equiv 0$) in different PDE families \cite{perko2013differential,fishwick2007handbook}. If the non-linear mapping $\zeta\left(\cdot,\cdot\right)$ is time-dependent, the accurate modelling of the long-term behaviours of a PDE (i.e., $\varepsilon\rightarrow\infty$) becomes increasingly challenging because the evolution rules of equation solution $\gamma\left(\cdot\right)$ change across time. To maintain accuracy, existing models are required to enlarge model size or complexity, inevitably facing the trade-off between accuracy and efficiency. Certainly, a model can constrain its strategy to only predict short-term dynamics (i.e., $\varepsilon=1$). However, learning long-term dynamics of PDEs serves as the cornerstone of diverse important applications, such as weather forecasting \cite{pathak2022fourcastnet}, epidemic prevention \cite{cao2020spectral}, economics analysis \cite{aminian2006forecasting}, and earth modelling \cite{xu2021spatiotemporal}. Therefore, overcoming the limitation in learning long-term PDE behaviours is necessary for optimizing PDE solvers, which is the core objective of our research. 

\subsection{Our framework and contribution} In this paper, we attempt to overcome the challenge of predicting long-term PDE dynamics by proposing a new neural operator named as the Koopman neural operator (KNO). Our framework and contributions are summarized as following:
\begin{itemize}
    \item[(1) ] \textbf{Long-term behaviour of the PDE family as a non-linear dynamic system of equation solution.} Besides learning the solution operator of an entire target PDE family, we formalize a non-linear dynamic system of equation solution described by Eq. (\ref{EQ6}) in the meanwhile. This characterization supports to optimize the iterative update strategy of neural operators in Eq. (\ref{EQ5}) using dynamic system theory.
    \item[(2) ] \textbf{Equivalent linear prediction of non-linear dynamics via Koopman operator.} To capture the intricate long-term dynamics, our model is designed to learn the Koopman operator, an infinite-dimensional linear operator governing all observations of a dynamic system, to act on the evolution mapping $\zeta\left(\cdot,\cdot\right)$ of the dynamic system of equation solution. By doing so, we can transform the original task into an equivalent but simpler linear prediction problem.
    \item[(3) ] \textbf{Balance between accuracy and efficiency in zero-shot and long-term prediction.} We compare the KNO with existing state-of-the-art models (e.g. the Fourier neural operator \cite{li2020fourier} and other competitors) in zero-shot prediction (i.e., testing on an untrained discretization granularity or an untrained prediction interval) and long-term prediction experiments on five representative PDEs (e.g., the Navier-Stokes equation and the Rayleigh-B{\'e}nard convection) and three real dynamic systems (e.g., global water vapor patterns and western boundary currents). In these experiments, the KNO is shown as a competitive approach to realize PDE solving and dynamic system modelling.
\end{itemize}
The open source code release is provided in \url{https://github.com/Koopman-Laboratory/KoopmanLab}. One can also see Ref. \cite{KoopmanLab2023} for this toolbox.

\section{The Koopman neural operator: theory and computation}\label{Sec2}

\subsection{Time-dependent Koopman operator}\label{Sec2-1}
The non-linear and potentially non-autonomous dynamic system in Eqs. (\ref{EQ6}-\ref{EQ7}) makes the long-term prediction a daunting challenge. In practice, researchers always expect to deal with a linear dynamic system (i.e., $\partial_{t}\gamma_{t}=\mathcal{A}\gamma_{t}$ where $\mathcal{A}$ is a linear operator), whose dynamics is sufficiently simple and can be accurately modelled. Therefore, a natural question is whether we can transform the original system in Eqs. (\ref{EQ6}-\ref{EQ7}) to a linear one to simplify the learning task with reasonable errors. 

According to modern dynamic system theory, such an objective can be realized by formulating the Koopman operator $\mathcal{K}$, an infinite-dimensional linear operator governing all possible observations of a dynamic system, and letting it act on the flow mapping $\zeta\left(\cdot,\cdot\right)$ to linearize the original dynamic system in an appropriate space \cite{brunton2022modern}. For instance, let us take the case where system $\gamma_{t}$ is autonomous (i.e., $\theta_{t}^{t^{\prime}}$ can be simplified as $\theta^{t^{\prime}}$) as a simple illustration, the family of Koopman operators $\mathcal{K}^{\varepsilon}:\mathcal{G}\left(\mathbb{R}^{d_{\gamma}}\times T\right)\rightarrow\mathcal{G}\left(\mathbb{R}^{d_{\gamma}}\times T\right)$, parameterized by time difference $\varepsilon$, is defined based on a set of observation function (or named as measurement function) $\mathcal{G}\left(\mathbb{R}^{d_{\gamma}}\times T\right)=\{\mathbf{g}\vert \mathbf{g}:\mathbb{R}^{d_{\gamma}}\times T\rightarrow\mathbb{C}\}$ \cite{brunton2022modern}
\begin{align}
\mathcal{K}^{\varepsilon}\mathbf{g}\left(\gamma_{t}\right)=\mathbf{g}\left(\theta^{\varepsilon}\left(\gamma_{t}\right)\right)=\mathbf{g}\left(\gamma_{t+\varepsilon}\right),\;\forall \gamma_{t}\in\mathbb{R}^{d_{\gamma}}\times T.\label{EQ9}
\end{align}
Given with an appropriate space defined by $\mathcal{G}\left(\mathbb{R}^{d_{\gamma}}\times T\right)$, we can linearize the dynamics of $\gamma_{t}$ via Eq. (\ref{EQ9}). This idea has seen notable success in fluid dynamics \cite{rowley2009spectral}, robotics \cite{abraham2019active}, plasma physics \cite{taylor2018dynamic}, and neuroscience \cite{brunton2016extracting}.

Different from existing machine-learning-based Koopman operator models that either are limited to autonomous dynamic systems (e.g., the case described by Eq. (\ref{EQ9})) \cite{takeishi2017learning,azencot2020forecasting,otto2019linearly,alford2022deep} or require \emph{a priori} knowledge about the eigenvalue spectrum (e.g, the numbers of real and complex eigenvalues) of Koopman operator for non-autonomous dynamic systems \cite{lusch2018deep}, our framework concerns a more general case where we consider a time-dependent Koopman operator applicable to both non-autonomous and autonomous dynamic systems \cite{macesic2018koopman}
\begin{align}
\mathcal{K}^{t+\varepsilon}_{t}\mathbf{g}\left(\gamma_{t}\right)=\mathbf{g}\left(\theta^{t+\varepsilon}_{t}\left(\gamma_{t}\right)\right)=\mathbf{g}\left(\gamma_{t+\varepsilon}\right),\;\forall t\leq t+\varepsilon\in T.\label{EQ10}
\end{align}
As shown in Eq. (\ref{EQ10}), this Koopman operator governs a time-dependent linear evolution flow of $\mathbf{g}\left(\gamma_{t}\right)$ in a space defined by $\mathcal{G}\left(\mathbb{R}^{d_{\gamma}}\times T\right)$
\begin{align}
\partial_{t}\mathbf{g}\left(\gamma_{t}\right)=\lim_{\varepsilon\rightarrow0}\frac{\mathcal{K}^{t+\varepsilon}_{t}\mathbf{g}\left(\gamma_{t}\right)-\mathbf{g}\left(\gamma_{t}\right)}{\varepsilon}.\label{EQ11}
\end{align}

In mathematics, the adjoint of the Koopman operator defined by Eq. (\ref{EQ9}) is the Perron-Frobenius operator of dynamic systems \cite{lasota1985probabilistic} while the adjoint of the associated Lie operator (see \ref{ASC-1} for details) is the Liouville operator of Hamiltonian dynamics \cite{gaspard1995spectral,gaspard2005chaos}. These properties relate our approach with well-known theories about linear representation of dynamic systems in statistical physics and quantum mechanics \cite{lasota1998chaos}.

\subsection{Time-dependent Koopman operator in PDE solving}\label{Sec2-2}

To understand how the linearization of $\mathbf{g}\left(\gamma_{t}\right)$ realized by the Koopman operator is related to PDE solving, we can consider the Lax pair $\left(\mathcal{M},\mathcal{N}\right)$ of an integrable version of the PDE described by Eqs. (\ref{EQ1}-\ref{EQ3}) \cite{lax1968integrals}
\begin{align}
\mathcal{M}&=\mathsf{D}_{x}^{n}+\alpha\gamma\left(x_{t}\right)I,\;\alpha\in\mathbb{C},\label{EQ12}\\
\mathcal{M}\psi\left(x_{t}\right)&=\lambda\psi\left(x_{t}\right),\;\lambda\in\mathbb{C},\label{EQ13}\\
\partial_{t}\psi\left(x_{t}\right)&=\mathcal{N}\psi\left(x_{t}\right),\label{EQ14}
\end{align}
in which $\mathsf{D}_{x}^{n}$ is the $n$-th total derivative operator and $I$ is an identity operator. We notice that Eq. (\ref{EQ13}) actually defines an eigenvalue problem associated with operator $\mathcal{M}$ at moment $t$. By calculating the time derivative of Eq. (\ref{EQ13}), we can observe a relation between linear operators $\mathcal{M}$ and $\mathcal{N}$
\begin{align}
\left(\partial_{t}\mathcal{M}+\mathcal{M}\mathcal{N}-\mathcal{N}\mathcal{M}\right)\psi\left(x_{t}\right)=\partial_{t}\lambda\psi\left(x_{t}\right).\label{EQ15}
\end{align}
This relation implies 
\begin{align}
\partial_{t}\mathcal{M}+\left[\mathcal{M},\mathcal{N}\right]=0,\label{EQ16}
\end{align}
where $\left[\mathcal{M},\mathcal{N}\right]=\mathcal{M}\mathcal{N}-\mathcal{N}\mathcal{M}$ stands for the commutator of operators \cite{lax2007linear}. Given Eqs. (\ref{EQ12}-\ref{EQ16}), we can combine them with Eq. (\ref{EQ11}) to identify a relation between operator $\mathcal{N}$ and the time-dependent Koopman operator $\mathcal{K}^{t+\varepsilon}_{t}$
\begin{align}
\psi\left(D\times\{t\}\right)=\mathbf{g}\left(\gamma_{t}\right)\Rightarrow\mathcal{N}=\lim_{t+\varepsilon\rightarrow t}\frac{\mathcal{K}^{t+\varepsilon}_{t}\mathbf{g}\left(\gamma_{t}\right)-\mathbf{g}\left(\gamma_{t}\right)}{\varepsilon}.\label{EQ17}
\end{align}
Note that Eq. (\ref{EQ17}) still holds when the evolution of equation solution $\gamma_{t}$ is autonomous (i.e., we can consider a time-invariant Koopman operator $\mathcal{K}$ in Eq. (\ref{EQ17}) directly). 

Therefore, the linearization of $\mathbf{g}\left(\gamma_{t}\right)$ via a Koopman operator is closely related to the Lax pair and can be understood in the aspect of the inverse scattering transform of integrable PDEs \cite{lax1968integrals}. This relation has been comprehensively studied in mathematics and physics \cite{parker2020koopman,nakao2020spectral,gin2021deep,page2018koopman}, ensuring the validity and effectiveness of using the Koopman operator theory during PDE solving.

\subsection{Computational idea of the Koopman operator approximation}\label{Sec2-3}
After formalizing the time-dependent Koopman operator and confirming its relation to PDE solving, our next task is explore a possible idea to computationally represent this Koopman operator. Inspired by the Hankel-DMD \cite{arbabi2017ergodic}, sHankel-DMD \cite{vcrnjaric2020koopman}, and HAVOK \cite{brunton2017chaos} algorithms, we consider the Krylov sequence \cite{saad2011numerical} of the observable defined by a unit time step $\varepsilon\in\left[0,\infty\right]$
\begin{widetext}
    \begin{align}
\mathcal{R}_{n}&=\left[\mathbf{g}\left(\gamma_{0}\right),\mathbf{g}\left(\gamma_{\varepsilon}\right),\mathbf{g}\left(\gamma_{2\varepsilon}\right),\ldots,\mathbf{g}\left(\gamma_{n\varepsilon}\right)\right],\label{EQ18}\\&=\left[\mathbf{g}\left(\gamma_{0}\right),\mathcal{K}^{\varepsilon}_{0}\mathbf{g}\left(\gamma_{0}\right),\mathcal{K}^{2\varepsilon}_{\varepsilon}\mathcal{K}^{\varepsilon}_{0}\mathbf{g}\left(\gamma_{0}\right),\ldots,\mathcal{K}^{n\varepsilon}_{\left(n-1\right)\varepsilon}\mathcal{K}^{\left(n-1\right)\varepsilon}_{\left(n-2\right)\varepsilon}\cdots\mathcal{K}^{\varepsilon}_{0}\mathbf{g}\left(\gamma_{0}\right)\right],\label{EQ19}
\end{align}
\end{widetext}
which can be seen in the Krylov subspace method for computing matrix eigenvalues \cite{saad2011numerical}. Such a sequence can be efficiently sampled by a Hankel matrix representation of the system
\begin{align}
\mathcal{H}_{m\times n}&=\begin{bmatrix} 
	\mathbf{g}\left(\gamma_{0}\right) & \mathbf{g}\left(\gamma_{\varepsilon}\right) & \cdots & \mathbf{g}\left(\gamma_{n\varepsilon}\right) \\
	\vdots & \vdots & \vdots & \vdots\\
	\mathbf{g}\left(\gamma_{\left(m-1\right)\varepsilon}\right) & \mathbf{g}\left(\gamma_{m\varepsilon}\right) & \cdots & \mathbf{g}\left(\gamma_{\left(m+n-1\right)\varepsilon}\right) \\
	\end{bmatrix},\label{EQ20}
\end{align}
whose dimension of delay-embedding is $m\in\mathbb{N}^{+}$. The columns of $\mathcal{H}_{m\times n}$ approximate the functions in the Krylov subspace. Our motivation to consider this Krylov subspace lies in the possibilities for it to span an invariant subspace, $\mathbb{K}\subset \mathcal{G}\left(\mathbb{R}^{d_{\gamma}}\times T\right)$, of the Koopman operator 
\begin{align}
\mathbb{K}=\operatorname{span}\left(\mathcal{R}_{n}\right)\simeq\operatorname{span}\left(\mathcal{H}_{\left(m,n\right)}\right)\label{EQ21}
\end{align}
if $n\geq \operatorname{dim}\left(\mathbb{K}\right)-1$ (notion $\operatorname{dim}\left(\cdot\right)$ denotes the dimensionality). To see these possibilities, we consider the Galerkin projection of the original Koopman operator to $\mathbb{K}$, which is denoted by $\widehat{\mathcal{K}}_{t}^{t+\varepsilon}:\mathcal{G}\left(\mathbb{R}^{d_{\gamma}}\times T\right)\rightarrow\mathbb{K}$ for any $t\in T$. For any function $\mathbf{h}\left(\cdot\right)\in\mathcal{G}\left(\mathbb{R}^{d_{\gamma}}\times T\right)$, we have
\begin{align}
\langle\widehat{\mathcal{K}}_{t}^{t+\varepsilon}\mathbf{h}\left(\gamma_{t}\right),\mathbf{g}\left(\gamma_{i\varepsilon}\right)\rangle=\langle\mathcal{K}_{t}^{t+\varepsilon}\mathbf{h}\left(\gamma_{t}\right),\mathbf{g}\left(\gamma_{i\varepsilon}\right)\rangle,\;\forall i=0,\ldots,m,\label{EQ22}
\end{align}
where $\langle\cdot,\cdot\rangle$ is the inner product. According to Refs. \cite{korda2018convergence,li2022reduced}, the Koopman operator restricted to $\mathbb{K}$ approximates the original Koopman operator
\begin{widetext}
    \begin{align}
\lim_{m\rightarrow\infty}\int_{\mathcal{G}\left(\mathbb{R}^{d_{\gamma}}\times T\right)}\Vert \widehat{\mathcal{K}}_{t}^{t+\varepsilon}\mathbf{h}\left(\gamma_{t}\right)-\mathcal{K}_{t}^{t+\varepsilon}\mathbf{h}\left(\gamma_{t}\right)\Vert_{F}\mathsf{d}\mu=0,\;\forall \mathbf{h}\left(\cdot\right)\in\mathcal{G}\left(\mathbb{R}^{d_{\gamma}}\times T\right) \label{EQ23}
\end{align}
\end{widetext}

if the original Koopman operator is bounded and $\mathbb{K}$ happens to be its invariant subspace (or simply $n\rightarrow\infty$). Notion $\mu$ denotes a measure on $\mathcal{G}\left(\mathbb{R}^{d_{\gamma}}\times T\right)$ and $\Vert\cdot\Vert_{F}$ is the Frobenius norm. Therefore, we are expected to approximate the original Koopman operator via a restricted counterpart such that 
\begin{align}
    \mathcal{H}_{m\times n}\left(k+1\right)=\widehat{\mathcal{K}}_{k\varepsilon}^{\left(k+1\right)\varepsilon}\mathcal{H}_{m\times n}\left(k\right),\;\forall k=1,\ldots,n, \label{EQ24}
\end{align}
where $\mathcal{H}_{m\times n}\left(k\right)$ denotes the $k$-th column of $\mathcal{H}_{m\times n}$. 

A challenge lies in that all the derivations presented above (i.e., the Krylov subspace method, Hankel matrix representation, and the Galerkin projection) are initially proposed for autonomous systems \cite{brunton2022modern,arbabi2017ergodic,vcrnjaric2020koopman,brunton2017chaos}. Although we can formulate these mathematical expressions for a time-dependent Koopman operator as shown in Eqs. (\ref{EQ18}-\ref{EQ24}), this formulation inevitably requires an online optimization in computational implementation because the concerned Koopman operator may change across time. The expensive online optimization is not favorable for solving PDEs, persuading us to consider an alternative solution. As suggested in Ref. \cite{li2023data}, we can assume a constant form of the Koopmen operator, $\widehat{\mathcal{K}}_{t}^{t+\varepsilon}$, only when $\varepsilon$ is sufficiently small (i.e., the time interval is small enough such that the changes of $\widehat{\mathcal{K}}_{t}^{t+\varepsilon}$ during this duration are negligible). Given this assumption, we can further consider a condition under which the dynamic system of $\mathbf{g}\left(\gamma_{t}\right)$ is approximately ergodic (i.e., the observation of equation solution, $\mathbf{g}\left(\gamma_{t}\right)$, eventually visits all possible states in $\mathbb{R}^{d_{\gamma}}$ as $t\rightarrow \infty$, thus the proportion of time that $\mathbf{g}\left(\gamma_{t}\right)$ spends on a particular state equals the probability of this state) \cite{arbabi2017ergodic,cornfeld2012ergodic}. This condition makes the time-averaging approximate the true expectation of of an observable as the time approaches to infinity. Under this condition, we can define an expectation of the Koopman operator controlled by time difference $\varepsilon$
\begin{widetext}
    \begin{align}
\overline{\mathcal{K}}_{\varepsilon}&=\lim_{t\rightarrow\infty}\frac{1}{t}\int_{\left[0,t\right)}\mathbf{g}\left(\gamma_{\tau}\right)^{-1}\mathbf{g}\left(\gamma_{\tau+\varepsilon}\right)\mathsf{d}\tau\simeq\operatorname*{argmin}_{P\in\mathbb{R}^{d_{\gamma}+1}}\sum_{k=1}^{n}\Vert\mathcal{H}_{m\times n}\left(k+1\right)-P\mathcal{H}_{m\times n}\left(k\right)\Vert_{F}.\label{EQ25}
\end{align}
\end{widetext}
Given a fixed $\varepsilon$, the Koopman operator $\overline{\mathcal{K}}_{\varepsilon}:\mathcal{G}\left(\mathbb{R}^{d_{\gamma}}\times T\right)\rightarrow\mathbb{K}$ in Eq. (\ref{EQ25}) can be understood as the time-average of $\widehat{\mathcal{K}}_{t}^{t+\varepsilon}$ at different $t$. A representation of $\overline{\mathcal{K}}_{\varepsilon}$ only requires offline optimization and is computationally favorable for solving PDEs.

Therefore, the offline optimization of the Koopman operator is required to support a high time resolution (i.e., a small $\varepsilon$). Meanwhile, we need to find an appropriate design of the observation function $\mathbf{g}\left(\cdot\right)$ such that the ergodicity condition holds with reasonable errors. Although these two requirements make analytic derivations or algorithmic approaches highly difficult (e.g., see Ref. \cite{li2023data} for the usage of an integrated framework of offline low rank decomposition and online reduced-order modeling in dealing with the first requirement), we suggest the possibility for a neural-network-based approach to satisfy them in practice.

\begin{figure*}[t!]
\includegraphics[width=1\columnwidth]{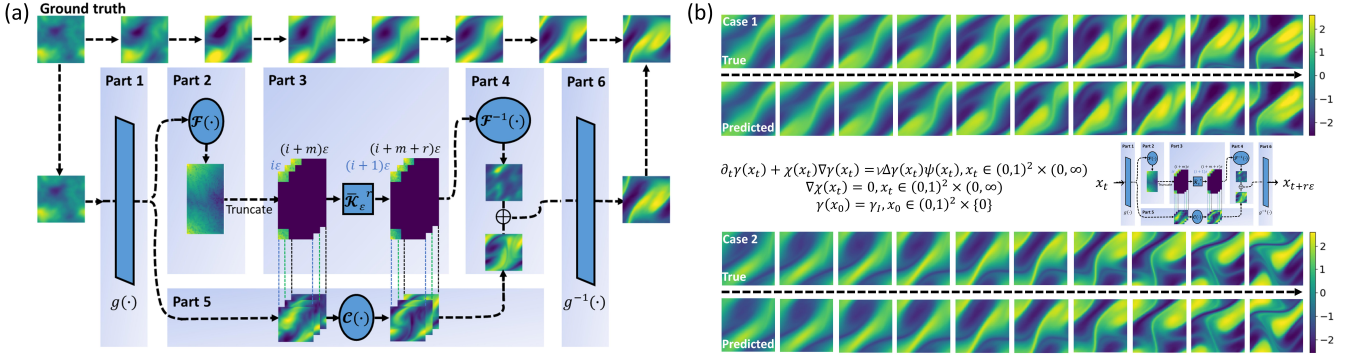}
\caption{\label{G0}Conceptual illustrations of neural network architectures of the KNO. Note that the layout of every part is slightly reorganized to offer a clear version.} 
 \end{figure*}

\subsection{Neural network architectures of Koopman neural operator}\label{Sec2-4}
The non-trivial part of the idea discussed above is how to define an effective neural network architecture to realize a Koopman-operator-based PDE solving pipeline. Below, we introduce the Koopman neural operator (KNO) as a possible architecture design, whose open-source PyTorch toolbox is released in Ref. \cite{xiong2023koopmanlab}. In general, a KNO architecture consists of the following parts:
\begin{itemize}
    \item[(1) ] \textbf{Part 1: Observation.} Given an input $\phi_{t}=\phi\left(D\times\{t\}\right)$ of the PDE in Eqs. (\ref{EQ1}-\ref{EQ3}), we first transform it as $\mathbf{g}\left(\widehat{\gamma_{t}}\right)$ in space $\mathcal{G}\left(\mathbb{R}^{d_{\widehat{\gamma_{t}}}}\times T\right)$ by an encoder (a single non-linear layer with $\tanh\left(\cdot\right)$ activation function) that represent an observation function $\mathbf{g}\left(\cdot\right)$. Please see Figure \ref{G0}for illustrations.
    \item[(2) ] \textbf{Part 2: Fourier transform.} We apply the Fourier transform to map $\mathbf{g}\left(\widehat{\gamma_{t}}\right)$ as $\mathbf{g}_{\mathcal{F}}\left(\widehat{\gamma_{t}}\right)=\mathcal{F}\circ\mathbf{g}\left(\widehat{\gamma_{t}}\right)$ and parameterize the subsequent parts of our network in the Fourier space. Similar to Ref. \cite{li2020fourier}, $\mathbf{g}_{\mathcal{F}}\left(\widehat{\gamma_{t}}\right)$ is computed by fast Fourier transform, where we truncate the Fourier series at $\omega$, a maximum number of frequency modes. On the one hand, as suggested in Ref. \cite{li2020fourier}, this procedure offers a convenient computation of the iterative update strategy shown in Eq. (\ref{EQ4}). On the other hand, the truncated Fourier transform is important for our model because it subdivides the system into two parts. The first part corresponds to the remaining low-frequency modes after truncation while the second part consists of the truncated high-frequency modes. Although there is no theoretical guarantee, low-frequency modes are generally more stable than those volatile high-frequency modes in practice unless the system is purely random. In other words, it is more possible for low-frequency modes to satisfy the ergodicity condition with acceptable errors than high-frequency modes. Therefore, by filtering out high-frequency modes and learning the Koopman operator only on low-frequency modes in \textbf{Part 3}, we can empirically reduce the difficulty of data-driven Koopman operator approximation. Meanwhile, we design an extra part (\textbf{Part 5}) in our model to extract the high-frequency modes of $\mathbf{g}\left(\widehat{\gamma_{t}}\right)$ to complement the lost information about high-frequency fluctuations. Thus, our model can also capture the volatile parts of real systems that are less ergodic and difficult to predict using the Koopman operator. Please see Figure \ref{G0}for details.
   \item[(3) ] \textbf{Part 3: Hankel representation and offline Koopman operator.} Given $\mathbf{g}_{\mathcal{F}}\left(\widehat{\gamma_{t}}\right)$ for every $t\in \varepsilon\mathbb{N}^{+}$, we set a dimension of delay-embedding, $m\in\mathbb{N}$, to define a Hankel matrix $\widehat{\mathcal{H}}_{m\times n}$ of $\mathbf{g}_{\mathcal{F}}\left(\widehat{\gamma_{t}}\right)$ (note that $n$ equals the number of accessible samples). To ensure that the space spanned by the Hankel matrix successfully approximates the invariant sub-space of target Koopman operator, we train a $o\times o$ linear layer to learn the $r$-th power of the Koopman operator ${\overline{\mathcal{K}}_{\varepsilon}}^{r}:\mathcal{G}\left(\mathbb{R}^{d_{\widehat{\gamma}}}\times T\right)\rightarrow\widehat{\mathbb{K}}$ following Eqs. (\ref{EQ24}-\ref{EQ25}). Based on the learned operator ${\overline{\mathcal{K}}_{\varepsilon}}^{r}$, we can predict the future state of the latest observable $\mathbf{g}_{\mathcal{F}}\left(\widehat{\gamma}_{\left(m+n-1\right)\varepsilon}\right)$ as $\mathbf{g}_{\mathcal{F}}\left(\widehat{\gamma}_{\left(m+n\right)\varepsilon}\right)=\left[{\overline{\mathcal{K}}_{\varepsilon}}^{r}\widehat{\mathcal{H}}_{m\times n}\left(n\right)\right]^{\mathsf{T}}\left(m\right)$, where notion $\mathsf{T}$ denotes the transpose of a matrix. Here we choose to learn the $r$-th power of $\overline{\mathcal{K}}_{\varepsilon}$ because a flexible value of $r\in\mathbb{N}$ offers opportunities for us to adjust the time resolution. Specifically, given the \emph{a priori} time resolution $\varepsilon$ pre-determined by the data set, we can control the internal time resolution of the learned Koopman operator by setting a prediction length $r$. Based on this setting, predicting the evolution of $\mathbf{g}_{\mathcal{F}}\left(\widehat{\gamma}_{t}\right)$ during a duration of $\varepsilon$ requires the Koopman operator to iterates $r$ times in the Fourier space (i.e., each iteration only corresponds to a short period of $\varepsilon/r$). In other words, even though the time resolution $\varepsilon$ of the sampled data may be not high enough, we can define a non-one value of $r$ to improve the actual time resolution on which the Koopman operator acts. This procedure has benefits in practice because the changes of the time-dependent Koopman operator are more negligible during a smaller period, which makes the offline optimization less challenging. Please see Figure \ref{G0}for illustrations of \textbf{Part 3}.
   \item[(4) ] \textbf{Part 4: Inverse Fourier transform.} Given each predicted state $\mathbf{g}_{\mathcal{F}}\left(\widehat{\gamma}_{\left(m+n\right)\varepsilon}\right)$ in \textbf{Part 3}, we transform it from the Fourier space to $\mathcal{G}\left(\mathbb{R}^{d_{\widehat{\gamma}}}\times T\right)$ by an inverse Fourier transform, i.e, $\mathbf{g}\left(\widehat{\gamma}_{\left(m+n\right)\varepsilon}\right)=\mathcal{F}^{-1}\circ\mathbf{g}_{\mathcal{F}}\left(\widehat{\gamma}_{\left(m+n\right)\varepsilon}\right)$. Note that high-frequency fluctuations in the system have been filtered out in \textbf{Part 2} and can not be directly recovered by the inverse Fourier transform (the complement of the lost information is realized by \textbf{Part 5}). Please see Figure \ref{G0}for the instances of \textbf{Part 4}.
   \item[(5) ] \textbf{Part 5: High-frequency information complement.} According to the Fourier analysis implemented on feature maps, convolutional layers can amplify high-frequency components \cite{park2022vision}. Therefore, we train a convolutional network $\mathcal{C}$ on the outputs of \textbf{Part 1} to extract their high-frequency information, denoted by $\mathbf{g}_{\mathcal{C}}\left(\widehat{\gamma_{t}}\right)$, as a complement of \textbf{Parts 2-4}. Meanwhile, the convolutional network also implements an independent forward prediction parallel to \textbf{Parts 2-4}, i.e, $\left[\mathbf{g}_{\mathcal{C}}\left(\widehat{\gamma}_{\left(i+1\right)\varepsilon}\right),\ldots,\mathbf{g}_{\mathcal{C}}\left(\widehat{\gamma}_{\left(i+m+1\right)\varepsilon}\right)\right]^{\mathsf{T}}=\mathcal{C}\left[\mathbf{g}\left(\widehat{\gamma}_{i\varepsilon}\right),\ldots,\mathbf{g}\left(\widehat{\gamma}_{\left(i+m\right)\varepsilon}\right)\right]^{\mathsf{T}}$ for any $i=1,\ldots,n$. In the released toolbox of the KNO \cite{xiong2023koopmanlab}, a basic implementation of $\mathcal{C}$ is a convolutional layer, which is simple and practical. If the forward prediction becomes challenging in complicated PDEs, one can also consider other effective convolutional network designs. For instance, we can use a simple tripartite network design to realize $\mathcal{C}$. The first and the third parts of $\mathcal{C}$ are convolutional layers used for data reshaping and the second part of $\mathcal{C}$ is the inception module (i.e., the basic component of the GoogLeNet) \cite{szegedy2015going}, whose efficiency has been extensively validated in practice. This design is provided in the toolbox as well. See Figure \ref{G0}for the illustrations of \textbf{Part 5}.
   \item[(6) ] \textbf{Part 6: Inverse observation.} Given two future states, $\mathbf{g}\left(\widehat{\gamma}_{\left(m+n\right)\varepsilon}\right)$ and $\mathbf{g}_{\mathcal{C}}\left(\widehat{\gamma}_{\left(m+n\right)\varepsilon}\right)$, of the latest observable independently predicted by \textbf{Parts 2-4} and \textbf{Part 5}, we unify them by $\mathbf{g}_{\mathcal{U}}\left(\widehat{\gamma}_{\left(m+n\right)\varepsilon}\right)=\left(1-\lambda\right)\mathbf{g}\left(\widehat{\gamma}_{\left(m+n\right)\varepsilon}\right)+\lambda\mathbf{g}_{\mathcal{C}}\left(\widehat{\gamma}_{\left(m+n\right)\varepsilon}\right)$, where $\lambda\in\left[0,1\right]$ controls the relative weights of low- and high-frequency information. We train an non-linear decoder (a single non-linear layer with $\tanh\left(\cdot\right)$ activation function) to represent the inverse of observation function $\mathbf{g}^{-1}\left(\cdot\right)\simeq\mathbf{g}_{\mathcal{U}}^{-1}\left(\cdot\right)$ and transform $\mathbf{g}_{\mathcal{U}}\left(\widehat{\gamma}_{\left(m+n\right)\varepsilon}\right)$ to $\widehat{\gamma}_{\left(m+n\right)\varepsilon}$, the target state of equation solution in space $\mathbb{R}^{d_{\widehat{\gamma}}}$. Please see Figure \ref{G0} for details.
\end{itemize}

Based on \textbf{Parts 1-6}, we have developed a new iterative update strategy different from Eq. (\ref{EQ4}). For any $t^{\prime}>t\in\varepsilon\mathbb{N}$, we have
\begin{widetext}
  \begin{align}
\widehat{\gamma}_{t^{\prime}}=\Big[\mathbf{g}^{-1}\Big(\underbrace{\mathcal{F}^{-1}\circ{\overline{\mathcal{K}}_{\varepsilon}}^{r}\circ\mathcal{F}\circ\mathbf{g}\left(\widehat{\gamma}_{\left[t-m\varepsilon,t\right]}\right)}_{\textbf{Parts 1-4}}+\underbrace{\mathcal{C}\circ\mathbf{g}\left(\widehat{\gamma}_{\left[t-m\varepsilon,t\right]}\right)}_{\textbf{Part 1 and part 5}}\Big)\Big]^{\mathsf{T}}\left(m\right),\label{EQ26}
\end{align}  
\end{widetext}
where $\widehat{\gamma}_{\left[t-m\varepsilon,t\right]}$ is a vector $\left[\widehat{\gamma}_{t-m\varepsilon},\ldots,\widehat{\gamma}_{t}\right]$ defined by $m\in\mathbb{N}$, the dimension of delay-embedding. In Figure \ref{G0}, we illustrate the one-unit architecture of KNO. Similar to Fourier neural operator \cite{li2020fourier}, a $x$-unit KNO architecture can be produced by cascading the copy of \textbf{Parts 2-5} $x$ times. Based on Eq. (\ref{EQ26}), the loss function of the KNO is defined as
\begin{align}
\mathcal{L}=\alpha\Vert \widehat{\gamma}_{t^{\prime}}-\gamma_{t^{\prime}}\Vert_{F}+\beta\sum_{i=0}^{m}\Vert\mathbf{g}^{-1}\circ\mathbf{g}\left(\widehat{\gamma}_{t-im\varepsilon}\right)-\gamma_{t-im\varepsilon}\Vert_{F},\label{EQ27}
\end{align}
where $\alpha,\beta\in\left(0,\infty\right)$ control the weights of prediction and reconstruction processes in the loss function, respectively. The idea of subdividing the prediction, reconstruction, low-frequency, and high-frequency contributions of the loss function can also be seen in previous studies about the sparse identification of nonlinear dynamic systems \cite{champion2019data,champion2019discovery,fasel2022ensemble}.

The proposed KNO framework shares a similar motivation of parameterizing the Koopman operator via neural networks with Ref. \cite{lusch2018deep}. The KNO is different from classic frameworks since it is developed for PDE solving and consists of multiple procedures to reduce the difficulties of Koopman operator approximation (e.g., the divide-and-conquer processing of high- and low-frequency information as well as the flexible setting of internal time resolution). Below, we test our KNO model in representative tasks to validate its effectiveness.

\section{Experiments}
\subsection{Experiments design}
\subsubsection{Data set information}
We implement our experiments on the 1-dimensional Bateman–Burgers equation \cite{benton1972table}, the 2-dimensional Navier-Stokes equation \cite{wang1991exact}, the 2-dimensional Rayleigh-B{\'e}nard convection (i.e., a kind of turbulent convection) \cite{bodenschatz2000recent}, the 2-dimensional shallow-water equations \cite{takamoto2022pdebench}, the infrared satellite imagery of water vapor in the Storm EVent ImagRy data set \cite{veillette2020sevir}, and the western boundary current data collected by the E.U. Copernicus Marine Environment Monitoring Service \cite{CopernicusMarine2021,hu2015pacific}. The first four data sets correspond to typical problems in turbulence analysis. The last two data sets contain the real data governed by unknown PDEs, which offer opportunities to explore the applicability of the KNO in real precipitation forecasting (e.g., storm and rainfall) and western boundary current analysis tasks (e.g., the dynamic modelling of Kuroshio and Gulf Stream currents). Please see \ref{ASC-2} for full data description. 

\subsubsection{Experiment designs and baselines}
We conduct five experiments to validate our model:
\begin{itemize}
    \item[(1) ] \textbf{Mesh-independent experiment.} As suggested in previous works \cite{lu2019deeponet,bhattacharya2020model,nelsen2021random,li2020neural,li2020fourier,kovachki2021universal,li2022fourier}, neural operator models are expected to be mesh-independent becasue they learn the solution operator of an entire PDE family. Therefore, we design an experiment to validate the mesh-independent property of the KNO.
    \item[(2) ] \textbf{Long-term prediction experiment.} To validate the long-term prediction capacity of the KNO, we design prediction tasks on eight data sets, including five representative PDE solving problems in turbulence analysis and three complicated dynamic system modelling problems.
     \item[(3) ]  \textbf{Zero-shot prediction experiment (discretization granularity).} Following the idea of Ref. \cite{li2020fourier}, we validate the generalization ability of the KNO by testing it on untrained discretization granularity (e.g., in a way similar to super-resolution \cite{li2020fourier}).
    \item[(4) ] \textbf{Zero-shot prediction experiment (prediction interval).} Apart from the generalization on untrained discretization granularity, we also validate the generalization capacity of the KNO on untrained prediction interval (e.g., zero-shot temporal interpolation and extrapolation).
    \item[(5) ] \textbf{Ablation experiment.} To demonstrate the significance of the learned Koopman operator in the KNO, we implement an ablation experiment.
\end{itemize}

In our experiments, the high-frequency complement can be either realized by a single convolutional layer, $\mathcal{C}_{s}$, or by a simple tripartite convolutional network, $\mathcal{C}_{t}$. Besides the KNO, we implement the following models for comparison: the Fourier neural operator (FNO) \cite{li2020fourier}, the U-shaped neural operator (UNO) \cite{rahman2022u}, the convolutional neural operator (CNO) \cite{raonic2024convolutional}, latent spectral model (LSM) \cite{wu2023solving}, the U-Net \cite{ronneberger2015u}, and the residual neural network (ResNet) \cite{he2016deep}. Among these baseline models, the FNO \cite{li2020fourier} is an extensively validated neural operator model. The UNO \cite{rahman2022u} and the CNO \cite{raonic2024convolutional} represent the performance of recently proposed neural operators. The LSM \cite{wu2023solving} serves as an instance of the neural network parameterization of spectral methods. The U-Net \cite{ronneberger2015u} and the ResNet \cite{he2016deep} represent the classic ideas of spatio-temporal modeling in the field of deep learning. Other common neural network models in deep learning, such as the convolutional LSTM (ConvLSTM) \cite{wang2020towards}, and deep hidden physics model (DHPM) \cite{raissi2018deep}, are shown as less efficient on complex dynamic systems \cite{wang2020towards} and are not considered in our research. Each model is trained by its default optimizer (e.g., the KNO and the FNO are trained by the Adam optimizer).

The experiments on the first four PDE data sets are implemented on a single Nvidia V100 GPU with 32GB memory. The experiments on the last two large-scale real data sets are implemented on a single Nvidia A100 GPU with 40GB memory.

\subsection{Experiment results}

\begin{figure*}[t!]
\includegraphics[width=1\columnwidth]{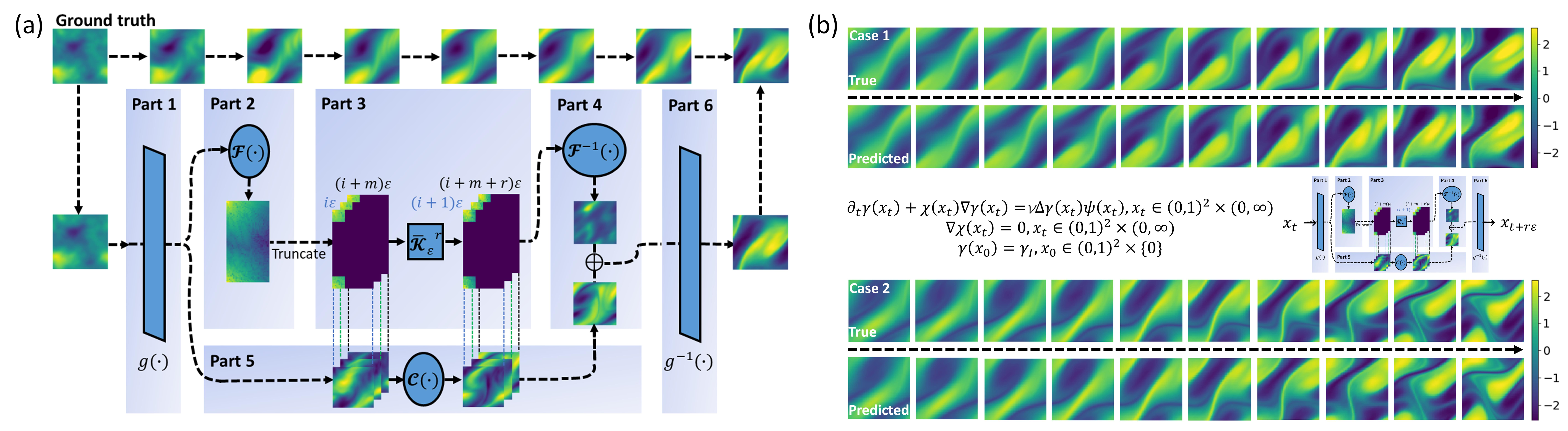}
\caption{\label{G1} Experiment results of mesh-independence. (a-d) respectively present the results under different conditions of high-frequency complement designs and $\lambda$ in the KNO. Here the high-frequency complement is either realized by a single convolutional layer, $\mathcal{C}_{s}$, or by a simple tripartite convolutional network, $\mathcal{C}_{t}$. Parameter $\lambda$ controls the contributions of low- and high-frequency information. Model performance is measured using the root mean square error (RMSE). In the legends of (a-d), the numbers within brackets indicate the parameter settings and model sizes, where the last number corresponds to the model size and all other numbers denote parameters. Notion $w$ in the FNO stands for the width parameter (i.e., the dimension of latent space) \cite{li2020fourier}. Note that the results of the FNO are not repeatedly shown in (b-d) since the adjustment of $\lambda$ has no relation with the FNO. Therefore, the performances of the KNO models in (b-d) can be directly compared with the performance of the FNO in (a).} 
 \end{figure*}

\begin{figure*}[t!]
\includegraphics[width=1\columnwidth]{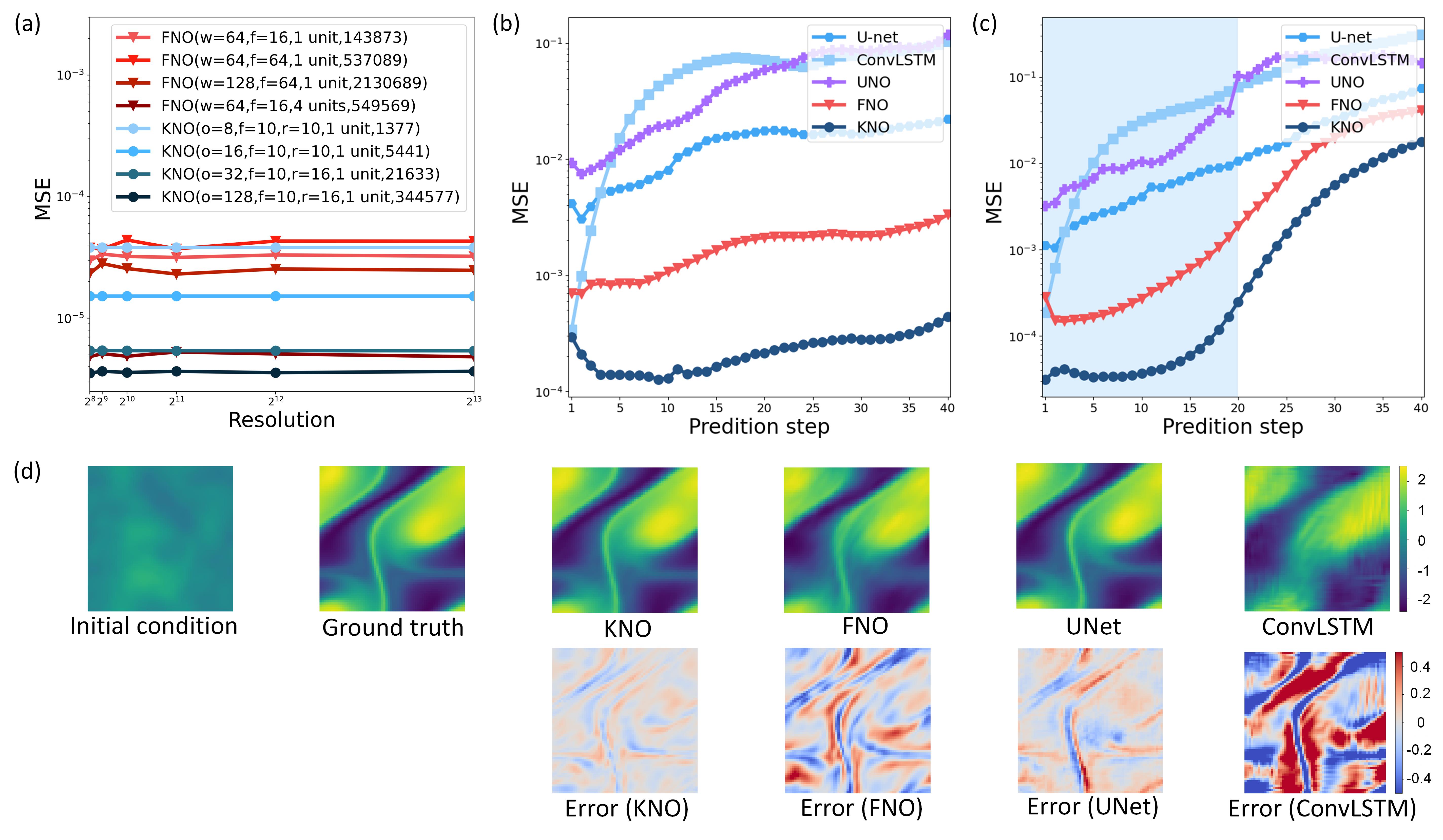}
\caption{\label{G2} Experiment results of long-term prediction. (a-h) show the performances of all models on different data sets, where each prediction step creates a time frame. (i) visualizes the prediction results (the twenty-sixth time frame after the initial condition is selected as an instance for illustration) and the associated errors of all models on the Rayleigh-B{\'e}nard convection. Note that color bars in (i) are shared by all models. Therefore, the results can be directly compared across different models. (j) presents the prediction results (the third time frame after the initial condition) and the associated errors of the KNO on the Kuroshio and Gulf Stream currents. } 
 \end{figure*}
 
\subsubsection{Mesh-independent experiment}

Our mesh-independent experiment is implemented on the data of 1-dimensional Bateman–Burgers equation generated under different discretization conditions (i.e., spatial resolution of meshes). The data with highest resolution is generated following the Gaussian initialization introduced in Ref. \cite{li2020fourier}. The data with lower resolution are directly down-sampled from the data with higher resolution. 

We choose the FNO as a baseline for comparison. As for other baseline models less efficient than FNO in mesh-independent experiment, such as graph neural operator (GNO) \cite{li2020neural} and multipole graph neural operator (MGNO) \cite{li2020multipole} (see results reported by Ref. \cite{li2020fourier}), we no longer discuss them for convenience. We implement multiple versions of the KNO with different hyper-parameters (e.g., operator size $o$, frequency mode number $f$, the iteration number $r$ of the Koopman operator, and the relative weight $\lambda$ that controls the contributions of low- and high-frequency information. Please see Sec. \ref{Sec2-4} for the meanings of these parameters). Under every condition, these models are trained on 1000 samples and conduct 1-second forward prediction (i.e., $t^{\prime}-t=1$) on 200 samples for performance evaluation. During training, the batch size is fixed as 64. The learning rate is initialized at 0.001 and is halved every 100 epochs. The weights of prediction and reconstruction in the loss function are defined as $\alpha=5$ and $\beta=0.5$, respectively. 

\begin{figure*}[t!]
\includegraphics[width=1\columnwidth]{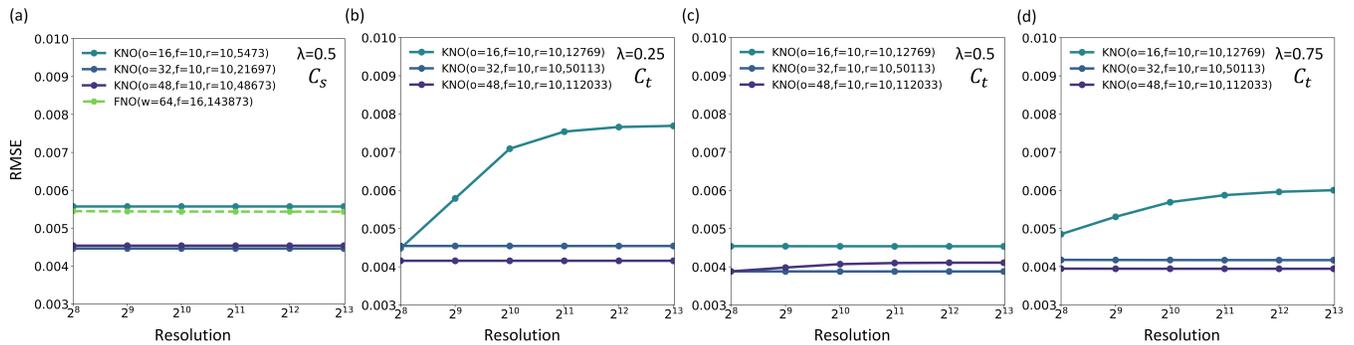}
\caption{\label{G3} Results of the zero-shot experiment (discretization granularity) on the 1-dimensional Bateman–Burgers equation. (a-d) respectively present the results under different conditions of high-frequency complement designs and $\lambda$ in the KNO. Here the high-frequency complement can be either realized by a single convolutional layer, $\mathcal{C}_{s}$, or by a simple tripartite convolutional network, $\mathcal{C}_{t}$. Parameter $\lambda$ controls the contributions of low- and high-frequency information. } 
 \end{figure*}

As illustrated in Figure \ref{G1}, the KNO achieves almost constant prediction error on every resolution. Compared with the FNO, the prediction error of KNO models are generally smaller under different conditions. Notably, a one-unit KNO only requires about $5\times 10^{3}$ parameters to achieve similar performance with a one-unit FNO that has more than $1.4\times 10^{5}$ parameters. When the model size of the KNO increases to $2\times 10^{4}$ or $4\times 10^{4}$, the KNO can significantly outperform the FNO. These results suggest the possibility for the KNO to be better at maintaining the balance between accuracy and efficiency in PDE solving (i.e., the KNO achieves higher accuracy with fewer parameters). Meanwhile, we notice that a KNO model with a balance between high- and low-frequency information (i.e., $\lambda=0.5$) can maintain more robust performance across different resolutions (i.e., the fluctuations of performance are relatively smaller). Therefore, we speculate that $\lambda=0.5$ is an optimal empirical choice in dealing with cross-resolution prediction tasks. This speculation is also verified in our subsequent zero-shot experiment in Figure \ref{G3}.

\subsubsection{Long-term prediction experiment}

Our long-term prediction experiment is implemented on all 2-dimensional data sets. Among these implemented data sets, the 2-dimensional Navier-Stokes equation can be defined with three viscosity cases, i.e., $\nu\in\{10^{-3},10^{-4},10^{-5}\}$. On all data sets, the trained models are required to predict $10$ time frames in the future (i.e., $t^{\prime}=t+z\varepsilon$, where $z\in\{1,\ldots,40\}$ is selected for the 2-dimensional Rayleigh-B{\'e}nard convection data while $z\in\{1,\ldots,10\}$ is set for all other data sets). Note that the meaning of the time difference $\varepsilon$ between adjacent time frames varies across different data sets. For the 2-dimensional Navier-Stokes equation, the 2-dimensional Rayleigh-B{\'e}nard convection, and the 2-dimensional shallow-water equations, the time difference denotes one second. For the the western boundary current data (e.g., the Kuroshio and Gulf Stream currents), the time difference corresponds to one day. For the infrared satellite imagery of water vapor, the time difference corresponds to a five-minute-interval. Therefore, ten or fifty time frames can already span a long interval of physics time to make the concerned systems exhibit long-term and highly non-linear evolution.

 Here we present the details of experiment implementation. For the Navier-Stokes equation, the training set contains 1000 samples. The testing set contains $200$ samples for $\nu\in\{10^{-3},10^{-4}\}$ while it includes $100$ samples for $\nu=10^{-5}$. For the Rayleigh-B{\'e}nard convection, models are trained and tested on $1600$ and $200$ samples, respectively. For the shallow-water equations, the training and testing sets respectively include $1600$ and $200$ samples. In all western boundary current data prediction tasks, models are trained on $1600$ samples and tested on $200$ samples. For the water vapor prediction task, models are tested on $400$ samples after being trained on $3200$ samples. To show that the KNO is not limited to small model designs, we present a possible realization of a relatively large KNO model (defined with $o=16$, $f=12$, $\lambda=0.5$, and $r=6$. High-frequency information complement is realized using the simple tripartite convolutional network $\mathcal{C}_{t}$. Model size is $19871979$). The compared models on all data sets include the FNO (model size is $926517$), the UNO (model size is $30478033$), the CNO (model size is $2667034$), the LSM (model size is $19188162$), the U-Net (model size is $24950491$), and the ResNet (model size is $20316490$). 

As shown in Figures \ref{G2}(a-h), the KNO achieves optimal or nearly optimal performance on all data sets. The increase rates of prediction errors in the KNO are generally smaller than those in other compared models. These results suggest that the usage of the Koopman operator modelling approach is beneficial to characterizing the long-term evolution of PDEs or real systems because it helps improves model accuracy. In Figures \ref{G2}(i-j), we show the instances of prediction results. It can be seen that the smaller errors achieved by the KNO do ensure a higher quality of 2-dimensional field visualization in its prediction results. The rapid growth of errors in other models generally lead to blurry and inaccurate visualization results.

\subsubsection{Zero-shot experiment on discretization granularity}
As suggested by Ref. \cite{li2020fourier}, a mesh-independent neural operator model can be trained only on the data with lower resolution and predict the data with higher resolution (referred to as zero-shot super-resolution \cite{li2020fourier}). In our research, we validate this property by implementing a zero-shot experiment and comparing between the KNO and the FNO.

Model settings are the same as our mesh-independence experiment shown in Figure \ref{G1}. We use the 1-dimensional Bateman–Burgers equation to implement the experiment. Most of the settings of prediction tasks keep the same as those in Figure \ref{G1}. The only difference lies in that all models are trained on a lower resolution and tested on a higher resolution. Specifically, models are trained on $2^{8}$ grids and tested on $\{2^{9},\ldots,2^{13}\}$ grids for the 1-dimensional Bateman–Burgers equation.

\begin{figure*}[t!]
\includegraphics[width=1\columnwidth]{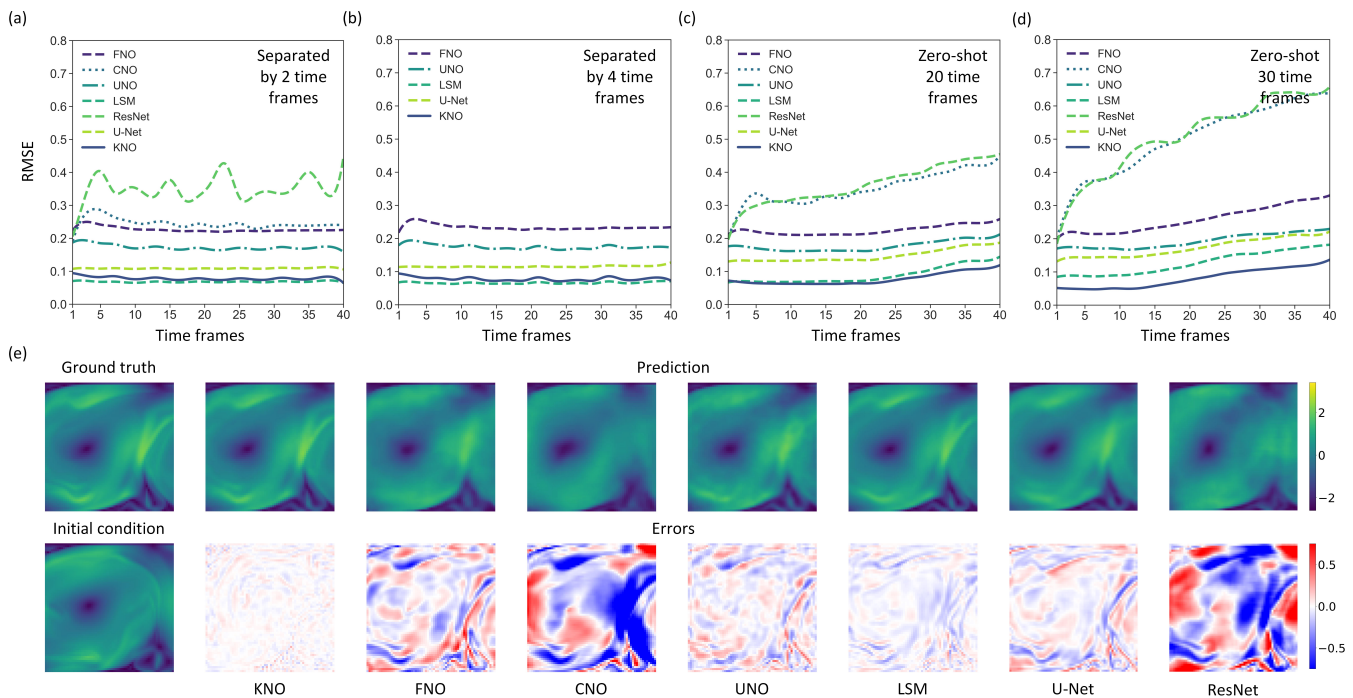}
\caption{\label{G4} Results of the zero-shot experiment (prediction interval) on the 2-dimensional Rayleigh-B{\'e}nard convection data. (a-b) show the results of the first kind of zero-shot prediction, where models are supervised by the time frames separated by 2 or 4 untrained frames during training, respectively. The result of the ResNet is absent in (b) because the ResNet does not converge well during training. (c-d) present the results of the second type of zero-shot prediction, where there are 20 or 30 unsupervised time frames during testing. Note that the lines shown in (a-d) are smoothed using the B-spline basis offered by Scipy \cite{virtanen2020scipy} because the raw performances of the ResNet and the CNO are oscillating. (e) visualizes the prediction results (the twenty-sixth time frame after the initial condition is selected as an instance for illustration) and the associated errors of all models during the zero-shot experiment shown in (d).} 
 \end{figure*}

As shown in Figure \ref{G3}, a KNO with a smaller model size can outperform a FNO with a larger model size under many conditions. Meanwhile, after comparing between the performances of the KNO across different settings of $\lambda$, we find that $\lambda=0.5$ is an optimal choice for zero-shot prediction across resolutions, which is consistent with our observations in Figure \ref{G1}.

\subsubsection{Zero-shot prediction experiment on prediction interval}

Because the KNO is proposed to model the intricate dynamics of PDE solutions, it is natural to wonder if the KNO can achieve optimal performance in the zero-shot prediction task with an untrained prediction interval. Specifically, there are two kinds of possible situations. In the first situation, models are only supervised by non-adjacent time frames separated by untrained time frames. Models are required to predict these unsupervised time frames separating between supervised ones. For instance, if supervised time frames are separated by two untrained frames, models learn how to predict the $1$-st, the $4$-th, the $7$-th time frames and so on during training. During testing, models are required to predict the $2$-nd, the $3$-rd, the $5$-th, and the $6$-th time frames and so on. In the second situation, models are supervised by a given time interval and learn how to predict the time frames within this interval. During testing, models are required to predict a much longer time interval that covers multiple unsupervised time frames. For instance, models are trained to predict 10 time frames in the future. During testing, they are required to predict 40 time frames instead. These two kinds of zero-shot experiments are useful in temporal interpolation and extrapolation.

The 2-dimensional Rayleigh-B{\'e}nard convection data is used to implement the experiment. We directly adopt model parameter settings from our experiment in Figure \ref{G2}. In the first kind of zero-shot experiment, models are supervised by the time frames separated by 2 or 4 untrained frames. In the second kind of zero-shot experiment, models are trained to predict 10 or 20 time frames. During the testing, they are required to predict 40 time frames in the future (i.e., there are 30 or 20 unsupervised time frames used for zero-shot prediction). As shown in Figures \ref{G4}(a-d), the KNO generally achieves ideal accuracy in both kinds of zero-shot experiments. In Figure \ref{G4}(e), we visualize instances of the prediction results of all models during the zero-shot extrapolation experiment with 30 unsupervised time frames. Because these instances correspond to the same system shown in Figure \ref{G2}(i), we can directly compare between these two cases and find that the long-term prediction of the KNO is still robust even without supervision while the performances of other models significantly reduce in the zero-shot case. These results suggest that the captured evolution patterns by the approximated Koopman operator are suitable for dynamic system modelling. 

\subsubsection{Ablation experiment}
To prove that the learned Koopman operator is a key factor to ensure the effectiveness of the KNO, we set a simple ablation experiment to compare the prediction performances between the KNO framework with and without the reconstruction loss term. In the former case, the temporal prediction in the model mainly depends on the weight matrix of the linear layer, which functions as the $r$-th power of a Koopman operator. Other optimizable parts of the model will be driven to learn the observation function and its inverse by the reconstruction loss. In the latter case, the whole model is driven to focus only on the temporal prediction and the linear layer does not necessarily function in a Koopman-operator-like manner.

We use the data of the 1-dimensional Bateman–Burgers equation to implement the ablation experiment. For convenience, the high-frequency complement in the KNO is realized by a single convolutional layer, $\mathcal{C}_{s}$, with $\lambda=0.5$. As shown in Table \ref{Tab0}, the KNO defined with an effective reconstruction process generally outperforms the KNO framework only driven by the prediction loss.

\begin{table*}
    \centering
\renewcommand\arraystretch{1}
    \begin{tabular}{lll|ll|l}
    \hline
      Operator size $o$ & Mode number $f$ & Iteration number $r$ & Weight $\alpha$ & Weight $\beta$ & RMSE \\
      \hline 
      8 & 10 & 10 &5.0 & 0.5 &  $6.18 \times10^{-3}$\\
      8 & 10 & 10 &5.0 & 0   &  \bm{$6.09\times10^{-3}$}\\
      \hline
      8 & 10 & 16 &5.0 & 0.5 & $5.01\times 10^{-3}$\\
      8 & 10 & 16 &5.0 & 0 & \bm{$4.79\times 10^{-3}$}\\
      \hline
      8 & 64 & 10 &5.0 & 0.5 &   \bm{$4.93\times 10^{-3}$}\\
      8 & 64 & 10 &5.0 & 0   &   $5.17 \times 10^{-3}$\\
      \hline
      16 & 10 & 16 &5.0 & 0.5 & \bm{$3.24\times 10^{-3}$}\\
      16 & 10 & 16 &5.0 & 0 & $3.32\times 10^{-3}$\\
      \hline
      32 & 10 & 16 &5.0 & 0.5 & \bm{$2.32\times 10^{-3}$}\\
      32 & 10 & 16 &5.0 & 0 & $2.46\times 10^{-3}$\\
      \hline
      32 & 64 & 16 &5.0 & 0.5 & \bm{$2.34\times 10^{-3}$}\\
      32 & 64 & 16 &5.0 & 0 & $2.40\times 10^{-3}$\\
      \hline
      128 & 10 & 16 &5.0 & 0.5 & \bm{$1.92\times 10^{-3}$}\\
      128 & 10 & 16 &5.0 & 0 & $2.08\times 10^{-3}$\\
      \hline
    \end{tabular}
        \caption{\label{Tab0}Results of the ablation experiment on the 1-dimensional Bateman–Burgers equation.}
\end{table*}

\section{Conclusion} 
In summary, we develop the Koopman neural operator (KNO), a mesh-independent neural-network-based solver of partial differential equations. The basic code of the KNO is provided in \ref{ASC-3} and the official toolbox of the KNO can be seen in Ref. \cite{KoopmanLab2023}. Compared with the existing state-of-the-art approaches in the field of neural-network-based PDE solving, such as the Fourier neural operator \cite{li2020fourier}, the U-shaped neural operator \cite{rahman2022u}, the convolutional neural operator \cite{raonic2024convolutional}, and latent spectral model \cite{wu2023solving}, our proposed KNO exhibits a higher capacity to capture the long-term evolution of PDEs or real dynamic systems. Meanwhile, it generally maintains accuracy across different mesh resolutions. This property suggests the potential of the KNO to be considered as a basic unit for constructing large-scale frameworks to solve complex physics equations with non-linear dynamics or predict future states of real dynamic systems. Moreover, the robust mesh-independence property enables the KNO to be trained on the low resolution data before being applied on the high resolution data. This property may create more possibilities to reduce the computational costs of PDE solving. 

\section*{Acknowledgements} Authors Y.T and P.S appreciate the supports from the Artificial and General Intelligence Research Program of Guo Qiang Research Institute at Tsinghua University (2020GQG1017) and the Huawei Innovation Research Program (TC20221109044). Author X.M.H is supported by the National Natural Science Foundation of China (42125503) and the National Key Research and Development Program of China (2022YFE0195900, 2021YFC3101600, 2020YFA0608000, 2020YFA0607900). Authors are grateful for the technical supports provided by Mr. Hao Wu, who studies at the School of Computer Science and Technology, University of Science and
Technology of China. Meanwhile, authors appreciate the helps of Mr. Yuan Gao and Mr. Shuyi Zhou, who study at Department of Earth System Science, Tsinghua University, during data preparation.


\appendix
\section{The associated Lie operator of the Koopman operator}\label{ASC-1}

Besides the linear system in Eq. (\ref{EQ9})) in the main text, we can also consider the generator operator of such a Koopman operator, which is referred to as the Lie operator because it is the Lie derivative of $\mathbf{g}\left(\cdot\right)$ along the vector field $\gamma\left(\cdot\right)$ \cite{koopman1931hamiltonian,abraham2012manifolds,chicone1999evolution}
\begin{align}
\mathcal{L}_{t}\mathbf{g}=\lim_{t+\varepsilon\rightarrow t}\frac{\mathcal{K}^{t+\varepsilon}_{t}\mathbf{g}\left(\gamma_{t}\right)-\mathbf{g}\left(\gamma_{t}\right)}{t+\varepsilon-t}.\label{EQA1}
\end{align}
The generator operator also defines a linear system of $\mathbf{g}\left(\gamma_{t}\right)$ because 
\begin{widetext}
    \begin{align}
\partial_{t}\mathbf{g}\left(\gamma_{t}\right)=\lim_{\varepsilon\rightarrow 0}\frac{\mathbf{g}\left(\gamma_{t+\varepsilon}\right)-\mathbf{g}\left(\gamma_{t}\right)}{\varepsilon}=\lim_{t+\varepsilon\rightarrow t}\frac{\mathcal{K}^{t+\varepsilon}_{t}\mathbf{g}\left(\gamma_{t}\right)-\mathbf{g}\left(\gamma_{t}\right)}{\varepsilon}=\mathcal{L}_{t}\mathbf{g}\left(\gamma_{t}\right).\label{EQA2}
\end{align}
\end{widetext}
Although our work primarily focuses on the Koopman operator, one can also model the Lie operator in application.

\section{Details of the implemented data sets}\label{ASC-2}
In our experiments, we consider the 1-dimensional Bateman–Burgers equation \cite{benton1972table}, the 2-dimensional Navier-Stokes equation \cite{wang1991exact}, the 2-dimensional Rayleigh-B{\'e}nard convection \cite{bodenschatz2000recent}, the 2-dimensional shallow-water equations \cite{takamoto2022pdebench}, the infrared satellite imagery of water vapor in the Storm EVent ImagRy data set \cite{veillette2020sevir}, and the western boundary current data collected by the E.U. Copernicus Marine Environment Monitoring Service \cite{CopernicusMarine2021,hu2015pacific}. Below, we introduce their precise definitions and pre-processing pipelines.

\subsection{Bateman–Burgers equation.} The 1-dimensional Bateman–Burgers equation is 
\begin{align}
    \partial_{t} u\left(x_{t}\right)+\partial_{x}\left(\frac{u^{2}\left(x_{t}\right)}{2}\right)&=\nu \partial_{xx} u\left(x_{t}\right),\;x_{t}\in\left(0,1\right)\times\left(0,1\right],\label{EQB1}\\
u\left(x_{0}\right)&=u_{I},\;x_{0}\in\left(0,1\right)\times\{0\},\label{EQB2}
\end{align}
where $u\left(\cdot\right)$ denotes the velocity, notion $\omega_{I}$ is a periodic initial condition, and parameter $\nu$ is the viscosity coefficient. The data set of Eqs. (\ref{EQB1}-\ref{EQB2}) is provided by Ref. \cite{li2020fourier}. The learning objective of this equation is $u\left(\cdot\right)$.

\subsection{Navier-Stokes equation.} Mathematically, the incompressible 2-dimensional Navier-Stokes equation in a vorticity form is defined as
\begin{align}
\partial_{t}\omega+\mathbf{u}\cdot\nabla\omega&=\nu\nabla^{2}\omega+\nabla\times\mathbf{f},\label{EQB3}\\
     \nabla\cdot\mathbf{u}&=0,\label{EQB4}
\end{align}
where $\omega$ is the scalar field of vorticity, notion $\mathbf{u}$ defines the vector field of velocity, and $\mathbf{f}$ denotes a time-independent forcing term. The viscosity coefficient is set as $\nu\in\{10^{-3},10^{-4},10^{-5}\}$. Similar to the situation of Bateman–Burgers equation, the data of Eqs. (\ref{EQB3}-\ref{EQB4}) is provided by Ref. \cite{li2020fourier}. In experiments, the learning target is vorticity field $\omega$.

\subsection{Rayleigh-B{\'e}nard convection equation.}
The 2-dimensional Rayleigh-B{\'e}nard convection is defined as \cite{bodenschatz2000recent}
\begin{align}
\partial_{t}\mathbf{u}+\mathbf{u}\cdot\nabla\mathbf{u}+\frac{1}{\rho\left(\theta\right)}\nabla p&=\nu\nabla^{2}\mathbf{u}+\mathbf{f},\label{EQB5}\\
\partial_{t}\theta+\mathbf{u}\cdot\nabla \theta&=\kappa\nabla^{2}\theta,\label{EQB6}\\
\nabla\cdot\mathbf{u}&=0,\label{EQB7}\\
\rho\left(\theta\right)&=\rho\left(\theta_{0}\right)\left[1-a\left(\theta-\theta_{0}\right)\right],\label{EQB8}
\end{align}
where $p$ is the pressure, scalar $\theta$ is the temperature, coefficient $\kappa$ measures the  heat conductivity, and $\rho\left(\theta\right)$ is the density of fluids given the temperature $\theta$. Notion $\theta_{0}$ is the initial temperature and $a$ denotes the  thermal expansion coefficient. The data of the 2-dimensional Rayleigh-B{\'e}nard convection is generated by Ref. \cite{wang2020towards} using the Boussinesq approximation approach. In our experiments, the learning target is the magnitude of velocity $\mathbf{u}$ (i.e., a scalar field).

\subsection{Shallow-water equations.}
The 2-dimensional shallow-water equations have the following forms \cite{takamoto2022pdebench}
\begin{align}
\partial_{t}h+\partial_{x}u+\partial_{y}v&=0,\label{EQB9}\\
\partial_{t}hu+\partial_{x}\left(u^{2}h+\frac{1}{g_{r}}h^{2}\right)&=-g_{r}h\partial_{x}b,\label{EQB10}\\
\partial_{t}hv+\partial_{y}\left(v^{2}h+\frac{1}{g_{r}}h^{2}\right)&=-g_{r}h\partial_{y}b,\label{EQB11}
\end{align}
where $h$ denotes a scalar field of water depth, notion $b$ describes the spatially varying bathymetry, and $g_{r}$ measures the gravitational acceleration. Notions $u$ and $v$ denote the velocities in horizontal and vertical directions, respectively. The data of the 2-dimensional shallow-water equations is offered by Ref. \cite{takamoto2022pdebench}. In our experiments, the learning objective is depth field $h$.

\subsection{Water vapor data.} The infrared satellite imagery of water vapor data is acquired from the Storm EVent ImagRy data set \cite{veillette2020sevir}, which can be used for precipitation analysis and forecasting (e.g., storm and rainfall). The spatial resolution of satellite imagery is $2$ kilometer and the time interval between each pair of adjacent imagery data is $5$ minutes. For convenience, we have down-sampled the data from $192\times 192$ to $128\times 128$ in the spatial domain.

\subsection{Western boundary current data.} The western boundary current data is acquired from the E.U. Copernicus Marine Environment Monitoring Service \cite{CopernicusMarine2021}. The data set can be used to analyze the dynamics of western boundary current, which is an important factor in shaping global ocean circulation and climate variability \cite{hu2015pacific}. Specifically, we use the the daily sea surface stream velocity data whose spatial resolution is $0.25\times0.25$ degrees. We choose the data in the Kuroshio region (10-42${}^{\circ}$N, 123-155${}^{\circ}$E) and the Gulf Stream region (20-52${}^{\circ}$N, 33-65${}^{\circ}$W) to study the Kuroshio and the Gulf Stream currents. The data covers the time interval from 2013/1/1 to 2018/12/31.

\section{Code implementation}\label{ASC-3}
The basic code implement of the KNO based on PyTorch is presented below. The official toolbox of the KNO is presented in \url{https://github.com/Koopman-Laboratory/KoopmanLab} (also see Ref. \cite{KoopmanLab2023}).

\begin{lstlisting}[language=Python]
import torch
import numpy as np
import torch.nn as nn
import torch.nn.functional as F

torch.manual_seed(0)

class encoder(nn.Module):
    def __init__(self, t_len, op_size):
        super(encoder, self).__init__()
        self.layer = nn.Linear(t_len, op_size)
    def forward(self, x):
        x = self.layer(x)
        x = torch.tanh(x)
        return x
    
class decoder(nn.Module):
    def __init__(self, t_len, op_size):
        super(decoder, self).__init__()
        self.layer = nn.Linear(op_size, t_len)
    def forward(self, x):
        x = torch.tanh(x)
        x = self.layer(x)
        return x

class Koopman_Operator1D(nn.Module):
    def __init__(self, op_size, modes_x = 16):
        super(Koopman_Operator1D, self).__init__()
        self.op_size = op_size
        self.scale = (1 / (op_size * op_size))
        self.modes_x = modes_x
        self.koopman_matrix = nn.Parameter(self.scale * torch.rand(op_size, op_size, self.modes_x, dtype=torch.cfloat))
    # Complex multiplication
    def time_marching(self, input, weights):
        # (batch, t, x), (t, t+1, x) -> (batch, t+1, x)
        return torch.einsum("btx,tfx->bfx", input, weights)
    def forward(self, x):
        batchsize = x.shape[0]
        # Fourier Transform
        x_ft = torch.fft.rfft(x)
        # Koopman Operator Time Marching
        out_ft = torch.zeros(x_ft.shape, dtype=torch.cfloat, device = x.device)
        out_ft[:, :, :self.modes_x] = self.time_marching(x_ft[:, :, :self.modes_x], self.koopman_matrix)
        #Inverse Fourier Transform
        x = torch.fft.irfft(out_ft, n=x.size(-1))
        return x
    
class KNO1d(nn.Module):
    def __init__(self, op_size, modes_x = 16, decompose = 4, t_len = 1):
        super(KNO1d, self).__init__()
        # Parameter
        self.op_size = op_size
        self.decompose = decompose
        # Layer Structure
        self.enc = encoder(t_len, op_size)
        self.dec = decoder(t_len, op_size)
        self.koopman_layer = Koopman_Operator1D(self.op_size, modes_x = modes_x)
        self.w0 = nn.Conv1d(op_size, op_size, 1)
    def forward(self, x):
        # Reconstruct
        x_reconstruct = self.enc(x)
        x_reconstruct = torch.tanh(x_reconstruct)
        x_reconstruct = self.dec(x_reconstruct)
        # Predict
        x = self.enc(x) # Encoder
        x = x.permute(0, 2, 1)
        x_w = x
        for i in range(self.decompose):
            x1 = self.koopman_layer(x) # Koopman Operator
            x = x + x1
        x = self.w0(x_w) + x
        x = x.permute(0, 2, 1)
        x = self.dec(x) # Decoder
        return x, x_reconstruct

class Koopman_Operator2D(nn.Module):
    def __init__(self, op_size, modes):
        super(Koopman_Operator2D, self).__init__()
        self.op_size = op_size
        self.scale = (1 / (op_size * op_size))
        self.modes_x = modes
        self.modes_y = modes
        self.koopman_matrix = nn.Parameter(self.scale * torch.rand(op_size, op_size, self.modes_x, self.modes_y, dtype=torch.cfloat))

    # Complex multiplication
    def time_marching(self, input, weights):
        # (batch, t, x,y ), (t, t+1, x,y) -> (batch, t+1, x,y)
        return torch.einsum("btxy,tfxy->bfxy", input, weights)

    def forward(self, x):
        batchsize = x.shape[0]
        # Fourier Transform
        x_ft = torch.fft.rfft2(x)
        # Koopman Operator Time Marching
        out_ft = torch.zeros(x_ft.shape, dtype=torch.cfloat, device = x.device)
        out_ft[:, :, :self.modes_x, :self.modes_y] = self.time_marching(x_ft[:, :, :self.modes_x, :self.modes_y], self.koopman_matrix)
        out_ft[:, :, -self.modes_x:, :self.modes_y] = self.time_marching(x_ft[:, :, -self.modes_x:, :self.modes_y], self.koopman_matrix)
        #Inverse Fourier Transform
        x = torch.fft.irfft2(out_ft, s=(x.size(-2), x.size(-1)))
        return x
    
class KNO2d(nn.Module):
    def __init__(self, op_size, modes = 10, decompose = 6, t_len = 10, weight = 0):
        super(KNO2d, self).__init__()
        # Parameter
        self.op_size = op_size
        self.decompose = decompose
        self.modes = modes
        # Layer Structure
        self.enc = encoder(t_len, op_size)
        self.dec = decoder(t_len, op_size)
        self.koopman_layer = Koopman_Operator2D(self.op_size, self.modes)
        self.weight = weight
        if self.weight == 0:
            self.w0 = nn.Conv2d(op_size, op_size, 1)
        else:
            self.w0 = Convolutional_Module(op_size) # Define a convolutional structure
    def forward(self, x):
        # Reconstruct
        x_reconstruct = self.enc(x)
        x_reconstruct = torch.tanh(x_reconstruct)
        x_reconstruct = self.dec(x_reconstruct)
        # Predict
        x = self.enc(x) # Encoder
        x = x.permute(0, 3, 1, 2)
        x_w = x
        for i in range(self.decompose):
            x1 = self.koopman_layer(x) # Koopman Operator
            x = x + x1
        x = (1 - self.weight) * x + self.weight * self.w0(x_w)
        x = x.permute(0, 2, 3, 1)
        x = self.dec(x) # Decoder
        return x, x_reconstruct
\end{lstlisting}

\bibliography{apssamp}
\end{document}